\pdfoutput=1
\documentclass[10pt,twocolumn,letterpaper]{article}

\usepackage{iccv}
\usepackage{times}
\usepackage{epsfig}
\usepackage{graphicx}
\usepackage{amsmath}
\usepackage{amssymb}
\usepackage{booktabs}
\usepackage{array}
\usepackage{dcolumn}
\usepackage{xcolor}

\usepackage[pagebackref=true,breaklinks=true,letterpaper=true,colorlinks,bookmarks=false]{hyperref}

\iccvfinalcopy 

\def\iccvPaperID{45} 

\ificcvfinal\fi

\begin{document}

\title{ Multi-Domain Few-Shot Learning and Dataset for Agricultural Applications}

\author{Sai Vidyaranya Nuthalapati\thanks{equal contribution; order determined by a coin toss}\\
Sensyne Health PLC\\
{\tt\small vidyaranya.ns@gmail.com}

\and
Anirudh Tunga\footnotemark[1]\\
Purdue University\\
{\tt\small atunga@purdue.edu}
}

\maketitle
\ificcvfinal\thispagestyle{empty}\fi

\begin{abstract}
   Automatic classification of pests and plants (both healthy and diseased) is of paramount importance in agriculture to improve yield. Conventional deep learning models based on convolutional neural networks require thousands of labeled examples per category. In this work we propose a method to learn from a few samples to automatically classify different pests, plants, and their diseases, using Few-Shot Learning (FSL). We learn a feature extractor to generate embeddings and then update the embeddings using Transformers. Using Mahalanobis distance, a class-covariance-based metric, we then calculate the similarity of the transformed embeddings with the embedding of the image to be classified. Using our proposed architecture, we conduct extensive experiments on multiple datasets showing the effectiveness of our proposed model. We conduct 42 experiments in total to comprehensively analyze the model and it achieves up to 14\% and 24\% performance gains on few-shot image classification benchmarks on two datasets. 
   
   We also compile a new FSL dataset containing images of healthy and diseased plants taken in real-world settings. Using our proposed architecture which has been shown to outperform several existing FSL architectures in agriculture, we provide strong baselines on our newly proposed dataset.
\end{abstract}

\section{Introduction}

In agriculture, correct classification of different pests and plants (both healthy and diseased) is a major issue due to the high similarity and shared characteristics between different species. Automatic classification of different categories using deep learning is an area of active research \cite{too2019deep,lu2017identification,ferentinos2018deep,xie2018multi}. However, in practice, people still rely on manual classification by experts for classification of different species. This can be partly attributed to the fact that using traditional deep learning networks based on Convolutional Neural Networks (CNNs) for classification require thousands of labelled examples per target category for training \cite{krizhevsky2012imagenet} and labeling samples on such a large scale requires domain experts. Further, the number of categories a trained CNN-based model can recognize remains fixed after training. To expand the set of categories that the network can recognize, it has to be further trained with new samples from novel classes, a process called fine-tuning \cite{chen2019closer}. Moreover, during training it is important that there is enough data (thousands per class) to prevent the network from overfitting \cite{gidaris2018dynamic}. Humans, on the other hand learn new tasks with very little supervision - a child can generalize the concept of ``horse" from a single picture. Also, humans can generalize to recognize novel categories from only one or a few examples \cite{lake2011one}. To enable networks to learn from a few examples, recently a lot of focus has been placed on Few-Shot Learning (FSL). FSL aims to tackle the problem of classification with very few training examples, and it is becoming popular in many fields \cite{chen2018a,li2006one,snell2017prototypical,yan2018few}. Specifically, in FSL, we are given two sets of labelled image data: \textit{meta-train} and \textit{meta-test} such that the image classes in both sets are mutually exclusive. The aim is to use the data in the meta-train set and learn transferable knowledge (also called meta-training) to construct a classifier on the visual classes in meta-test set which can classify a given \textit{query} sample even with very few labeled (\textit{support}) examples. Further, if the domain of visual classes in the meta-train set is different from that of meta-test set it is called cross-domain FSL. Feature distribution discrepancies across domains is one of the main challenges in cross-domain FSL. Similarly, in mixed-domain FSL, both the meta-train and meta-test sets contain classes from multiple domains and single-domain FSL contains instances from a single domain.

Recent FSL approaches in agriculture \cite{li2020few,li2021meta} learn an embedding function using the meta-train set which is then used to generate embeddings of the samples in the meta-test set. Finally, for a given query sample, a distance metric applied on the embeddings of the support samples gives the final prediction. A major limitation of this approach is the assumption that the transferable knowledge learned using meta-train classes can be directly applied to classification tasks generated using meta-test classes \cite{ye2020few}. For example, this approach assumes that the discriminative features for differentiating two plant species is same as the discriminative features required for differentiating two pest species. This problem becomes even more pronounced in cross-domain settings. In this work, we adopt an adaptation method that modifies the representations derived from the embedding function. The modified representations are tailored to maximize the discriminative power of the visual representations for a task at hand. Following \cite{ye2020few}, we use a Transformer \cite{vaswani2017attention} as a set-to-set function approximator that adapts the generated embeddings to the current task. 

Secondly, the existing few-shot architectures in agriculture \cite{argueso2020few, li2021meta} focus on using the Euclidean distance to compute the similarity of instance representations. However, using the Euclidean distance assumes that the feature dimensions are un-correlated and the feature dimensions have uniform variance \cite{bateni2020improved}, which do not always hold. Hence, inspired by \cite{bateni2020improved}, we use Mahalanobis distance \cite{galeano2015mahalanobis}, a  class-covariance-based metric, to compute the similarity of the embeddings.

Finally, the existing few-shot plant datasets for agriculture \cite{li2021meta, hughes2015open} contain images of a single leaf in laboratory settings as shown in Fig \ref{fig:plant}. However, in real-world applications of agriculture, we rarely get such images of a single leaf for classification. For instance, an image captured by a person using the ubiquitous smartphone camera is very different and contains challenging scenarios of lighting, orientation and background. As a result, it is imperative to design and test machine learning frameworks that perform well even with images containing the entire plants and not just a single leaf under varied conditions. To address this issue, we collected samples from publicly available resources (e.g., {\url{https://edenlibrary.ai/home}}) and compile a dataset of healthy and diseased plants. We also provide a strong baseline for our new few-shot dataset.

To test our architecture, we conduct extensive FSL experiments on multiple datasets, namely, the Plant and Pests (PP) dataset \cite{li2021meta} and the PlantVillage dataset \cite{hughes2015open}. The former is a dataset of plants and pests and enables us to experiment with mixed and cross-domain  settings. We improve the state-of-the-art accuracy of this dataset significantly, both in mixed-domain and cross-domain settings. On the other hand, the PlantVillage dataset consists of images of healthy and diseased plants. Our model outperforms the current best-performing models on this dataset by a significant margin.

The contributions of this work are three-fold: 1) We propose a new architecture for FSL using Transformers for enhancing the feature representations and class-covariance-based distance metric for calculating the feature distance. 2) We conduct extensive experiments under different settings on  multiple datasets to establish the superiority of the model. Various settings include single-domain, mixed-domain, and cross-domain (including cross-dataset). 3) We collect samples from publicly available resources to create a new FSL dataset containing images of healthy and diseased plants which represent the real-world applications of computer vision in agriculture. We also provide strong baselines on the proposed dataset.
\vspace{-1em}
\section{Related Work}
In the field of agriculture, many recent works on vision have tried to solve the classification task with limited training samples. In \cite{hu2019low}, the authors used  DC-GAN \cite{radford2015unsupervised} to generate augmented images for training.  Another approach proposed in  \cite{li2020few} is based on prototypical networks \cite{snell2017prototypical} and trains a CNN feature extractor followed by Euclidean distance calculation. The framework is trained using  a triplet loss function. Another recent work \cite{li2021meta}, used FSL to train a model with limited samples. In \cite{argueso2020few}, the authors trained a CNN to extract general plant leaf characteristics and used Siamese networks combined with triplet loss for classification. In \cite{argueso2020few}, the authors tested their work using the PlantVillage \cite{hughes2015open} dataset. In \cite{li2021meta}, the authors used a part of the PlantVillage \cite{hughes2015open} dataset. 

The existing FSL datasets \cite{hughes2015open,li2021meta,argueso2020few} of plants contain images of plants in an ideal setting with contrasting backgrounds, single leaves, no occlusions, and constant lighting conditions. To facilitate development of robust computer vision approaches in agriculture, we introduce \textit{Plants in Wild}, a new FSL dataset with images of diseased and healthy leaves in real-life setting.

In general, there are two main components of FSL: 1) learning generalizable instance embeddings and 2) distance computation between support and query instances to classify the input images.
In many works, generalizable instance embeddings \cite{hsu2018unsupervised,metz2018meta,vinyals2016matching,triantafillou2017few,changpinyo2017predicting} are learned and they are used for further classification using simple classifiers like nearest-neighbor methods and linear classifiers. In \cite{vinyals2016matching}, the authors use a nearest neighbor approach. MetaOptNet \cite{lee2019meta} uses a linear classifier. Siamese networks \cite{koch2015siamese} use a shared feature extractor and classification is done using the smallest L1 distance between the query sample and the support samples. Our work improves on the current models by altering both the components of FSL stated above. Following \cite{ye2020few}, we change the way instance embeddings are learnt by enhancing the  embeddings using a transformer \cite{vaswani2017attention} based set-to-set function approximator. Transformers have been effective to contextualize the representations of inputs and have found applications in diverse fields  \cite{kant2020spatially, tunga2020pose}. Secondly,  to compute the similarity between query and support examples, we use class-covariance based deterministic metric - Mahalanobis distance \cite{galeano2015mahalanobis, bateni2020improved}, which takes into account the distribution in feature space of each class, resulting in improved non-linear classifier decision boundaries.

\begin{figure*}
\begin{center}
\includegraphics[scale=0.40]{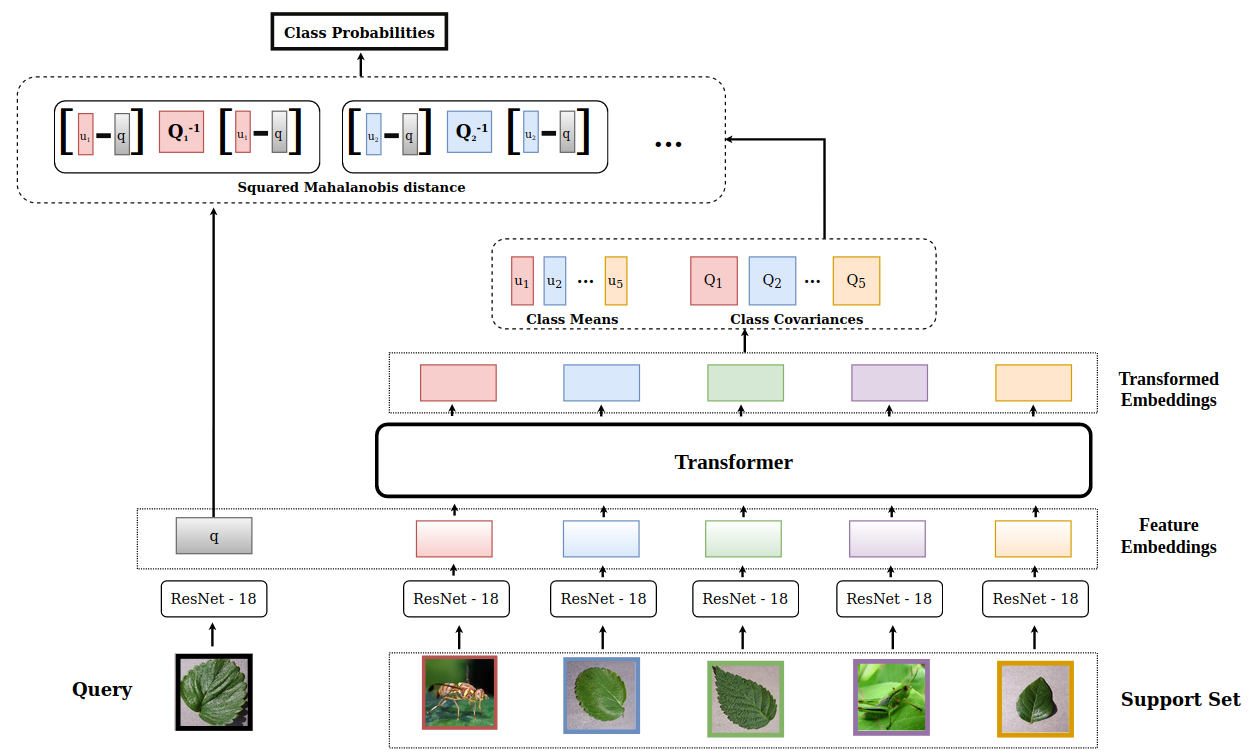}

\end{center}
  \caption{Illustration of the proposed architecture. We adapt the extracted features of the support set for each task using a Transformer before calculating the class-covariance based distance between support set features and query features. }
\label{fig:arch}
\end{figure*}

\section{Problem formulation}
Assume that we are given two labeled sets $\mathbb{D}_{meta\_train}$ and $\mathbb{D}_{meta\_test}$, such that the labels of the two sets are mutually exclusive i.e. $y_{meta\_train} \cap y_{meta\_test} = \phi$. In both the datasets $y_i$ is a one-hot encoded vector denoting the label of the corresponding sample.

In the FSL paradigm, we are given a set of episodes $\{T^{i}\}$. Each episode $T^{i}$ is made up of a support set and a query set. We denote support set by $\mathbb{D}_{support}$ and the examples in the support set by $(x_{s}, y_{s})$. Similarly, we denote the examples in a query set ($\mathbb{D}_{query}$) by $(x_{q}, y_{q})$. We also note that some existing works refer to support set, query set, and episode as training set, test set, and task respectively. In this work, we use these terms interchangeably. Each episode is represented as an $M$-shot $N$-way classification problem, where $N$ classes are randomly selected from a set of classes  $y_{meta\_test}$ with $M$ support examples for each of the $N$ classes. The goal is to learn a classifier \textit{f} such that given a query sample $x_{q}$ and the support set $\mathbb{D}_{support}$, the classifier is able to predict $y_{q}$. In a few-shot scenario the value of $M$ is very small (1, 5, 10).

In order to learn the parameters of the classifier \textit{f} we use a larger dataset called a meta-training dataset denoted by $\mathbb{D}_{meta\_train}$ to sample multiple $M$-shot $N$-way episodes \cite{vinyals2016matching, ye2020few, finn2017model}. In each of the episodes, the classifier labels the input $x_{q}^{meta\_train}$ as one of the \textit{N} $\in y_{meta\_train}$ classes. During the learning process, we minimize the average loss value of all the sampled tasks. To evaluate the performance of the classifier, we followed similar steps on $\mathbb{D}_{meta\_test}$.

\section{Architecture}
The classifier \textit{f} consists of two main components: an \textit{embedding function} and a \textit{distance calculation}.   Next,  we discuss both of these components in detail.

\begin{figure*}
\begin{center}
\includegraphics[scale=0.33]{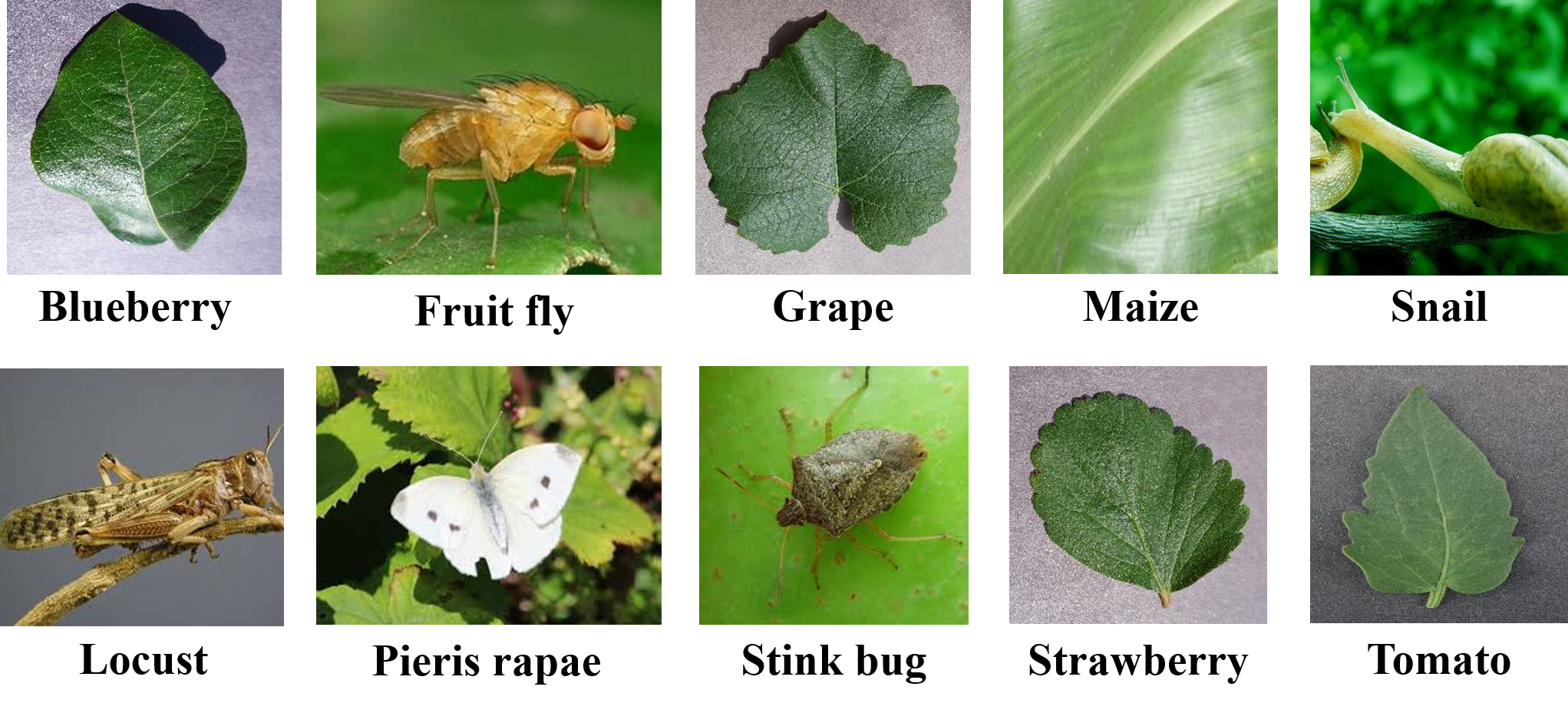}
\vspace{-10pt}
\end{center}
   \caption{An example of meta-train set (top row) and meta-test set (bottom row) for mixed domain classification on PP dataset \cite{li2021meta}.}
\label{fig:single}
\end{figure*}
\begin{figure*}
\begin{center}
\includegraphics[scale=0.33]{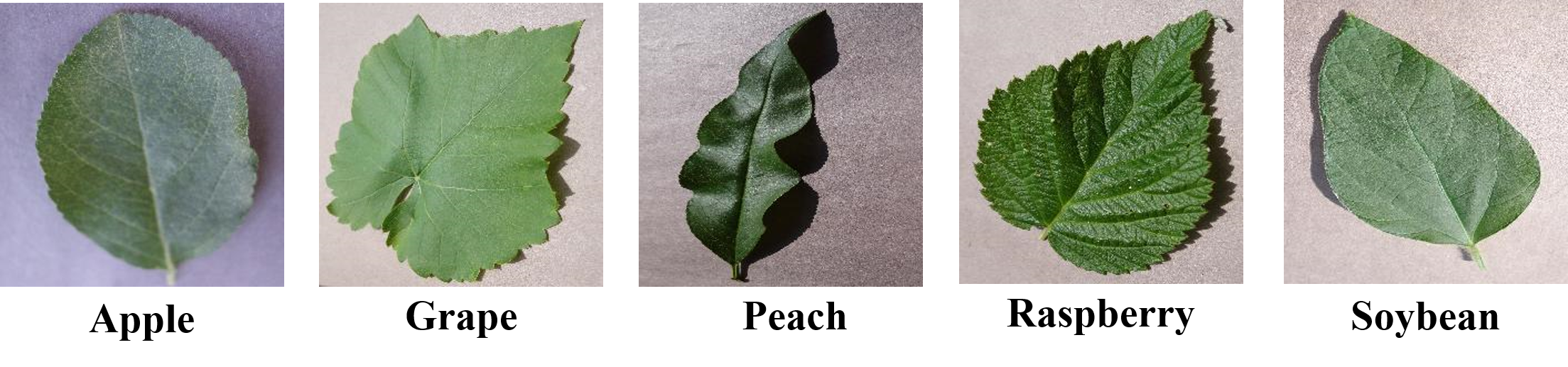}
\vspace{-10pt}
\end{center}
  \caption{An example of set containing samples only from plant classes from PP dataset \cite{li2021meta}.}
\label{fig:plant}
\end{figure*}

\subsection{Embedding function} The \textit{embedding function} is trainable and consists of two parts: feature extraction $\phi_{x}$ and transformation $\psi_{x}$. The feature extraction step $\phi_{x}$ extracts the features and projects them to a $d$-dimensional space. The feature extractor architecture consists of a ResNet-18 \cite{he2016deep} network pre-trained on Imagenet \cite{russakovsky2015imagenet}. The extracted features, however, are not ideal and do not capture important discriminative visual features for a specific episode\\task \cite{ye2020few}. In other words, simply using an embedding generated by $\phi_{x}$ does not incorporate any information about other support samples in the current episode. As a result, irrespective of the cohort, the embedding of a given image is always the same. Having such deterministic features is detrimental for discriminating the elements of the set. To overcome this, the embeddings need to be contextualized. Inspired by FEAT algorithm proposed in \cite{ye2020few}, the outputs of $\phi_{x}$ are transformed using a set-to-set function ($\psi_{x}$), which enriches the embedding of each image by considering those of all other images in the support set. By enabling interaction between embeddings of various images in the set it leads to richer representations and discriminative features. Another desired feature of the set-to-set function is that it should be permutation-invariant. Following \cite{ye2020few}, we implement $\psi_{x}$ using \textit{Transformer} architecture \cite{vaswani2017attention}, which fit the criteria of the required set-to-set function. Transformers rely on the self-attention mechanism which updates the representation of every input by weighing the relevance of all the other inputs. They map a query and a set of key-value pairs to an output.

\begin{equation}
   \psi_{x_{i}} =  \phi_{x_{i}} + \left(  \sum_{\forall j} V(s_j)f( \phi_{x_{i}} ,  \phi_{x_{j}} )  \right),
    \label{eq:first}
\end{equation}

where $f( \phi_{x_{i}} ,  \phi_{x_{j}} )$ is used to measure the similarity between $\phi_{x_{i}} ,\phi_{x_{j}} $.

\begin{equation}
    f( \phi_{x_{i}} ,  \phi_{x_{j}} ) = softmax_j\left(\frac {Q(\phi_{x_{i}} )^{T}K(\phi_{x_{j}} )} {\sqrt{d}}\right),
    \label{eq:second}
\end{equation}

 where the functions \textit{Q} and \textit{K} are learned linear projections. Combined with \textit{V}, which is also a learned linear projection, the functions \textit{Q} and \textit{K} project the inputs to a common representation space before applying the similarity measure. 

The resultant transformed embeddings $(\psi_{x})$ act as inputs to the distance computation module.

\vspace{5pt}
\begin{figure*}
\begin{center}
\includegraphics[scale=0.33]{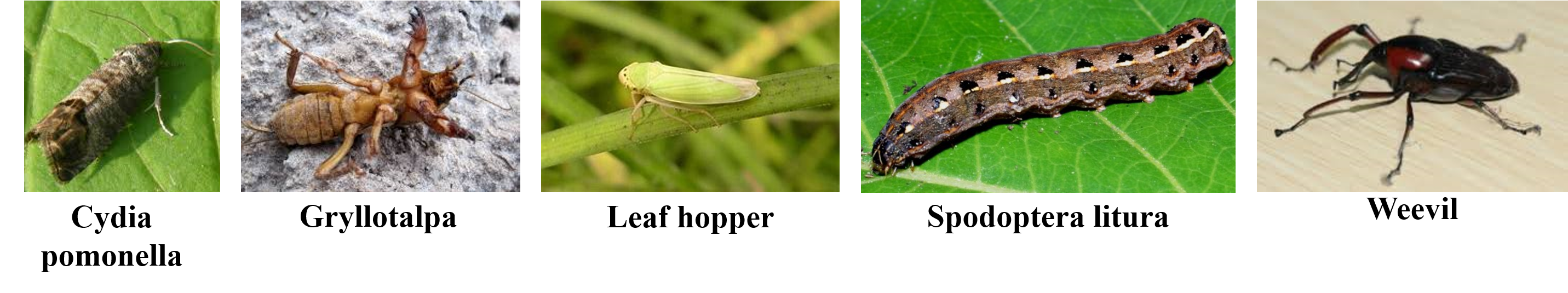}
\vspace{-10pt}
\end{center}
  \caption{An example of set containing samples only from pest classes from PP dataset \cite{li2021meta}.}
\label{fig:pest}
\end{figure*}

\subsection{Distance computation}
In this step, we calculate the distance between the embeddings of the support samples with the query sample to finally classify the query. Following \cite{bateni2020improved}, we use Mahalanobis distance \cite{galeano2015mahalanobis}, a class-covariance based distance metric estimated at test-time, which has been shown to be better than using other metrics such as Euclidean or cosine. The final prediction is calculated as follows:
\begin{equation}
    p(y_q = n | \psi_{x_q}, S^t) = \text{softmax}(-d_n(\psi_{x_q}, \mu_n)),
\end{equation}
where $n \in N$ is one of the support classes in the current episode/task $t \in T$, $S^t$ is the support set for the current episode/task, $\mu_n$ is the mean transformed embedding ($\psi_{x}$) of all the $M$ samples corresponding to the class $n$ and $d_n$ is the squared Mahalanobis distance defined as below:

\begin{equation}
    d_n(x, y) = \frac{1}{2}(x - y)^T(Q_{n}^{t})^{-1}(x - y),
\end{equation}
where $Q_{n}^{t}$ is the covariance matrix of class $n \in N$ for  episode/task $t \in T$. We calculate the covariance matrix $Q_{n}^{t}$ using a regularized estimator 

\begin{equation}
    Q_{n}^{t} = \lambda_{k}^{t} \Sigma_{n}^t + (1 - \lambda_{k}^{t}) \Sigma^t + \beta \textit{I},
\end{equation}
where $\Sigma^t$ is the covariance matrix of all the classes in the task $t$, $\Sigma_{n}^t$ is the covariance matrix of the class $n$ in task $t$ and $\lambda$ is a weigthing factor defined as follows:
\begin{equation}
\lambda_{k}^{t} = \frac{|S_k^t|}{|S_k^t| + 1},
\end{equation}

where $|S_k^t|$ is the number of elements in the support set of the task $t$ corresponding to class $k$. For a theoretical explanation of the superiority of Mahalanobis distance to other metrics, we refer the reader to \cite{bateni2020improved}.

\subsection{Loss function}
To ensure that the transformation function $\psi_{x}$ pulls the embeddings of same-class instances closer and drives different-class instances farther, we use a contrastive loss function. To achieve this, the transformation function is applied to instances of each of the $N$ classes present in the given episode/task, giving a transformed embedding $\psi_{x}'$ and the mean class centers $\{c_{n}\}_{n=1}^{N}$. We then calculate the similarity of individual embedding to the class center corresponding to the individual. Apart from this we also use a standard cross-entropy loss function to calcualte the loss using the final prediction. Hence, the complete loss function  is as follows:
\begin{equation}
\begin{split}
 \mathcal{L}(\hat{y}_{q}, & y_{q})  = l(\hat{y}_{q}, y_{q}) \\
   & + \lambda \times l(\text{softmax}(\text{sim}(\phi_{x}, c_{n})), y_{q}),
    \end{split}
\end{equation}
where $sim$ is the similarity function calculated using the class-covariance-based distance metric, $l$ is the standard cross entropy loss function and $\lambda$ is the weighting factor.

\subsection{Implementation Details}
The proposed model has been implemented using PyTorch \cite{paszke2019pytorch}. We use stochastic gradient descent with Nesterov acceleration with an initial learning rate of 0.0002, weight decay of 5e-4, momentum of 0.9, and a learning rate scheduler with step size of 40 and gamma of 0.5 to optimize the model. We resize all the images to 84×84×3 before using the feature extractor. The value of $\lambda$ is set to 0.1.
\begin{table}[h!]
    \caption{Various classes in our newly proposed \textit{Plants in Wild} dataset}
    \vspace{10pt}
    \label{tab:table1}
    \begin{tabular}{l|l} 
       {Healthy} & {Diseased}   \\
      \hline
      Celery & Corn Leaf Blight\\
      Chinese Cabbage & Chinese Cabbage Fusarium \\
      Cotton & Corn Rust Leaf\\
      Grapevine & Grapevine Esca\\
      Potato & Potato Early Blight \\
      Red Cabbage & Cucumber Tetranychus\\
      Tomato & Tomato Fruit Virus \\
      Watermelon & Cucumber Thrips \\
      Zucchini & Grapevine Powdery Mildew \\
      Bell Pepper & Bell Pepper Leaf Spot \\
    \end{tabular}
\end{table}

\section{Experiments}
In this section, we provide an overview of the datasets used, describe the experimental setup and provide quantitative results.

\begin{figure*}
\begin{center}
\includegraphics[scale=0.5]{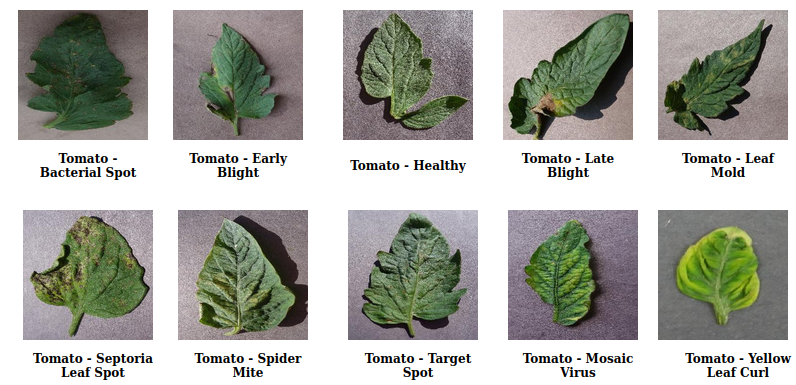}
\vspace{-10pt}
\end{center}
  \caption{Examples of healthy and diseased tomato leaves from PlantVillage dataset \cite{hughes2015open}. }
\label{fig:plantvillage}
\end{figure*} 

\vspace{5pt}
\begin{figure*}
\begin{center}
\includegraphics[scale=0.5]{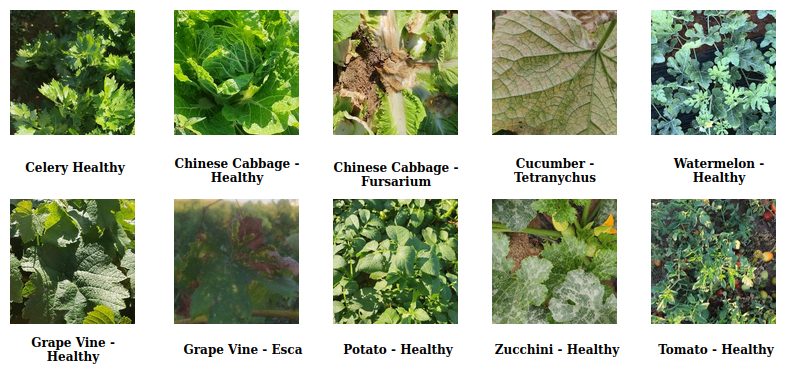}
\vspace{-10pt}
\end{center}
   \caption{A few examples of real world images from our Plants in Wild (PiW) dataset containing both healthy and diseased plants.}
\label{fig:ourdataset}
\end{figure*}

\subsection{Datasets}\label{sec:datasets}
We test our proposed architecture on two different challenging datasets and also provide baselines for the new dataset that we propose. We next describe the two external datasets that we use.

\textbf{Plant and Pest (PP)} \cite{li2021meta} dataset contains 6000 images of both pests and plants. We use this dataset to test the mixed-domain and cross-domain performance of our proposed architecture. The dataset contains 20 classes, 10 each for plants and pests. During experimentation, the dataset is split into two: meta-train set and meta-test set. The labels of both sets do not overlap and the distribution of the sets may also differ. In mixed domain settings, both meta-train and meta-test sets contain instances of plants and pests. We show an example of this in Fig \ref{fig:single}. On the other hand, there are two cross-domain settings. In the cross-domain 1 setting, the meta-train set contains examples of pests and the meta-test set is made up of plant instances. For example, Fig \ref{fig:pest} can make up the meta-train set and the meta-test set contains images from Fig \ref{fig:plant}. Finally, the cross-domain 2 setting, refers to the meta-train set containing plants and the meta-test set containing pest instances. In this setting, the images in Fig \ref{fig:plant} make up the meta-train set, whereas instances in Fig \ref{fig:pest} constitute the meta-test set. While being challenging, the cross-domain setting is vital owing to its practical implications. 

\textbf{PlantVillage} \cite{hughes2015open} is a public dataset containing 38 classes of healthy and diseased leaves of different crops. The dataset consists of 61,486 images. We split the dataset into 3 different and mutually exclusive meta-training and meta-testing sets, in the same manner followed in \cite{argueso2020few}. In split 1, the meta-test set consists of 10 different classes of Tomato (1 healthy, and 9  diseased) as shown in Fig \ref{fig:plantvillage}, and the meta-training set consists of the rest of the 28 classes. In split 2, the meta-test set consists of 4 classes of Apple (3 diseased, 1 healthy): apple scab, black rot, cedar apple rust, and healthy; 4 classes of Grape (3 diseased, 1 healthy): black rot, esca, leaf blight, and healthy; 2 classes of Cherry (1 diseased, 1 healthy): powdery mildew, and healthy; and the meta-train consists of rest of the 28 classes. Similarly, in split 3, the meta-test set consists of 4 classes of Corn (1 healthy, 3 diseased), 4 classes of Grape (1 healthy, 3 diseased), and 2 classes of peach (1 healthy, 1 diseased). 

\begin{table*}[ht!]
\caption{Average accuracy achieved on different setting of N and M on mixed domain on PP dataset \cite{li2021meta}.}
\vspace{2pt}
\centering
\begin{tabular}{lc cccccc}
\toprule
& \multicolumn{2}{c}{K = 1} & \multicolumn{2}{c}{K = 5} & \multicolumn{2}{c}{K = 10} \\
\cmidrule(r){2-3}\cmidrule(l){4-5}\cmidrule(l){6-7}
$N_{train}, N_{test}$ & {Li et al. \cite{li2021meta}} & Ours  & {Li et al. \cite{li2021meta}} & Ours   & {Li et al. \cite{li2021meta}}  & Ours   \\
\midrule
3, 3  & 0.811 & \textbf{0.841} & 0.870 & \textbf{0.900} & 0.904 & \textbf{0.929} \\
3, 5 & 0.677 & \textbf{0.723} & 0.794 &  \textbf{0.839}& 0.821 &\textbf{ 0.873} \\
5, 3 & 0.790 & \textbf{0.828} & 0.864 & \textbf{0.912} & 0.889 & \textbf{0.924} \\
5, 5& 0.676 & \textbf{0.739} & 0.780 & \textbf{0.837} & 0.825 & \textbf{0.872}\\

\bottomrule
\end{tabular}
\label{tab:single_mixed}
\end{table*}

\begin{table*}[ht!]
\caption{Average accuracy achieved on different settings of N and M on cross-domain 1 on PP dataset \cite{li2021meta}.}
\centering
\vspace{2pt}
\begin{tabular}{lc cccccc}
\toprule
& \multicolumn{2}{c}{M = 1} & \multicolumn{2}{c}{M = 5} & \multicolumn{2}{c}{M = 10} \\
\cmidrule(r){2-3}\cmidrule(l){4-5}\cmidrule(l){6-7}
$N_{train}, N_{test}$ & {Li et al. \cite{li2021meta}} & Ours    & {Li et al. \cite{li2021meta}} & Ours   & {Li et al. \cite{li2021meta}}  & Ours\\
\midrule
3, 3  & 0.697 & \textbf{0.792}  & 0.849 & \textbf{0.937} & 0.865 & \textbf{ 0.955}\\
3, 5 & 0.614 &\textbf{0.705} & 0.774 & \textbf{0.894} & 0.812 & \textbf{0.930 }\\
5, 3 & 0.721 & \textbf{0.759} & 0.814 &\textbf{ 0.923} & 0.871 & \textbf{0.936} \\
5, 5& 0.531 & \textbf{0.677} & 0.760 & \textbf{0.884} & 0.810 & \textbf{0.902}\\

\bottomrule
\end{tabular}
\label{tab:cross1}
\end{table*}

\begin{table*}[ht!]
\caption{Average accuracy achieved on different setting of N and M on cross-domain 2 on PP dataset \cite{li2021meta}}
\centering
\vspace{2pt}
\begin{tabular}{lc cccccc}
\toprule
& \multicolumn{2}{c}{M = 1} & \multicolumn{2}{c}{M = 5} & \multicolumn{2}{c}{M = 10} \\
\cmidrule(r){2-3}\cmidrule(l){4-5}\cmidrule(l){6-7}
$N_{train}, N_{test}$ & {Li et al. \cite{li2021meta}} & Ours  & {Li et al. \cite{li2021meta}} & Ours   & {Li et al. \cite{li2021meta}}  & Ours   \\
\midrule
3, 3  & 0.441 & \textbf{0.518} & 0.526 & \textbf{0.660} & 0.548 & \textbf{0.716} \\
3, 5 & 0.290 &\textbf{0.381} & 0.380 & \textbf{0.547} & 0.427 &  \textbf{0.594}\\
5, 3 & 0.425 & \textbf{0.520} & 0.531 & \textbf{0.665} & 0.558 & \textbf{0.708} \\
5, 5& 0.292 & \textbf{0.385} & 0.374 & \textbf{0.545} & 0.439 & \textbf{0.603}\\

\bottomrule
\end{tabular}
\label{tab:cross2}
\vspace{-1em}
\end{table*}

\textbf{Plants in Wild (PiW)}: In addition to these two datasets, we curated another dataset to validate our model's performance across domains and under real-life settings. The images of leaves in the Plantvillage and PP dataset are in a laboratory setting such that each image contains only one leaf in a contrasting background. However, in a real-life setting, the images are less likely to be taken in a controlled setting and will be of varying backgrounds, lighting conditions, and angles. We used the publicly available images available on the internet (for instance, the Eden Library website\footnote{\url{https://edenlibrary.ai/home}}) to collect images of diseased and healthy plants. All the images were taken using smartphone cameras in different fields and farms. We preprocessed all the images by resizing and center-cropping the images. Fig \ref{fig:ourdataset} shows some of the classes from our dataset. Comparing our dataset with the existing datasets, shown in Figures \ref{fig:plant} and \ref{fig:plantvillage}, highlights the differences between the datasets - lab controlled and real-life images. 

We collected 20 categories of plants: 10 of which are healthy and 10 classes correspond to diseased plants. The dataset consists of 1980 images, which were taken by users using different smartphone cameras.  Table \ref{tab:table1} shows all the available classes in the dataset. We evaluate the model using three different splits of the dataset. In split 1, the meta-train set consists of Bell pepper healthy, Bell pepper leaf spot, Celery healthy, Chinese cabbage healthy, Chinese cabbage fusarium, Corn leaf blight, Corn rust leaf, Cucumber tetranychus, Cucumber thrips, and Red cabbage healthy. The meta-test set consists of the rest of the 10 classes. In split 2, the meta-test and meta-train sets are reversed. In split 3, the meta-train set consists of all the healthy plant leaf images, while the meta-test set consists of all the diseased plant leaf images.

\subsection{Results}
We evaluate the performance of our proposed model on the PP dataset \cite{li2021meta} and the PlantVillage dataset \cite{hughes2015open} to establish the superiority of our model compared to existing best-performing methods on these datasets. We then use our model and provide baselines for our newly proposed dataset. To evaluate the model, we generate 600 support-and-query sets from the meta-test set. To report the final accuracy, we use the average accuracy on all the 600 sets.


\begin{table*}[ht!]
\caption{Average accuracy achieved on different settings on PlantVillage dataset \cite{hughes2015open}}
\centering
\begin{tabular}{lc cccccc}
\toprule
& \multicolumn{2}{c}{Split 1} & \multicolumn{2}{c}{Split 2} & \multicolumn{2}{c}{Split 3} \\
\cmidrule(r){2-3}\cmidrule(l){4-5}\cmidrule(l){6-7}
Method used & 5 way 1 shot  & 5 way 5 shot   & 5 way 1 shot  & 5 way 5 shot &  5 way 1 shot  & 5 way 5 shot  \\
\midrule
Argueso et al. \cite{argueso2020few}  & 0.34 & 0.531 & 0.464 & 0.769 & 0.552 & 0.693 \\
\textbf{Ours - Cross Dataset} & \textbf{0.38} & \textbf{0.54} & \textbf{0.573} & \textbf{0.772} & \textbf{0.716} & \textbf{0.861}\\
\textbf{Ours} & \textbf{0.466} & \textbf{0.635} & \textbf{0.709} & \textbf{0.87} & \textbf{0.754} & \textbf{0.885}\\

\bottomrule
\end{tabular}
\label{tab:plantvillage}
\vspace{-1.5em}
\end{table*}


\textbf{Plant and Pest (PP)}: On the \textit{PP dataset} \cite{hughes2015open}, we perform 36 sets of experiments, under three different settings: 2 cross-domain and mixed-domain following \cite{li2021meta}. Please refer to the \ref{sec:datasets} for the details on different settings. 
For each of the domains we evaluate with different values of \textit{M} and \textit{N}. We also vary the value of \textit{N} during the meta-train and meta-test phase, which we represent using $N_{meta\_train}$ and $N_{meta\_test}$. For example, a value of 3, 5 corresponding to $N_{meta\_train}$ and $N_{meta\_test}$ respectively means that the number of support classes in each episode is 3 during meta-training and 5 during the model evaluation (or the meta-testing) phase. We also experiment with 3 different values of M (1, 3, and 5). We observe significant performance gains across all the settings. Specifically, we observe around 3 to 6\% gains in performance on the mixed-domain setting, approximately 7 to 11\% improvement on cross-domain 1, and about 12 to 14 \% improvement on cross-domain 2 setting. We observe that gains on cross-domain settings are much higher compared to the mixed domain setting. This shows that our architecture is much better at extracting discriminative features even on unseen domains. We also observe that the accuracy improvements when using a higher number of support classes during meta-testing are greater i.e. when using a value of 5 for $N_{meta\_test}$ compared to 3. Specifically, the average improvement across all the settings with a $N_{meta\_test}$ value of 5 is 9\% while it is equal to 5\% when $N_{meta\_test}$ is set to 3.  This is due to the fact that, with lower values of $N_{meta\_test}$ even a random classifier can correctly predict the outcome with high probability (given just 3 three classes, a random classifier is correct 33\% of the time). Increasing the value demands robust architectures for accurate predictions. In summary, we see the trend of the increasing performance gap with increasing difficulty of the task throughout our experiments: performance improvements on cross-domain settings are more pronounced compared to single-domain and the accuracy gap widens when we increase the number of support classes in meta-test set.

\textbf{PlantVillage}: On the PlantVillage dataset \cite{hughes2015open}, we perform experiments under 6 different settings. We use three different splits (please refer to the Section \ref{sec:datasets} for details on splits) and use two different values of $M$ (1, 5) while keeping $N$ constant. Please refer to the row corresponding to \textit{Ours} in Table \ref{tab:plantvillage} for the results. Experiments using our model outperform the current state-of-the-art architecture \cite{argueso2020few} on this dataset by 10 to 24\%. Moreover, the  performance gains are much higher in 1-shot settings with a mean improvement of 18\% compared to an average improvement of 11\% in 5-shot settings. This is in line with our observations on the PP dataset \cite{li2021meta} that the performance gains are much higher as the tasks get tougher.

\textbf{Plants in Wild (PiW)}: The PiW dataset, in contrast to the previous two datasets, consists of diseased and healthy images of plants taken in a natural setting. We perform six experiments using three different data splits (Please refer to the \ref{sec:datasets} for the details on different splits) to validate the performance of our model in a natural setting. As shown in Table \ref{tab:piwres}, our model performs the same, if not better when compared to its performance on plant images in a controlled setting. These experiments show the effectiveness of our model even with real-life images taken with smartphone cameras.  

\textbf{Cross Dataset Performance:} To further establish the superiority of our model in adapting to unseen domains, we only use the images of pests from the PP dataset \cite{li2021meta} as our meta-train set and use plant images in the PlantVillage dataset \cite{hughes2015open} as meta-test set. We show that even under such a stringent \textit{cross-domain / cross-dataset} setting, our algorithm still shows notable improvements (up to 17\% gains). We record these values in the \textit{Ours - Cross Dataset} row in Table \ref{tab:plantvillage}.

In summary, results across multiple settings on various datasets indicate that using Transformers for generating episode-specific rich instance embeddings ($\psi_{x}$) provide for a useful calculation of episode and class specific covariance matrix ($Q_{n}^{t}$) in few-shot settings.

\begin{table}[h] 
\centering
\caption{Average accuracy on different settings on our PiW dataset.}
\vspace{2pt}
\begin{tabular}{cccc}
\hline\noalign{\smallskip}
Splits & N,M  & \textbf{Ours}  \\
\noalign{\smallskip}\hline\noalign{\smallskip}
Split 1 & 
\begin{tabular}{c}5 way 1 shot \\5 way 5 shot\end{tabular}
& \begin{tabular}{c}0.768 \\0.905\end{tabular}\\
\noalign{\smallskip}\hline\noalign{\smallskip}
Split 2 & 
\begin{tabular}{c}5 way 1 shot \\5 way 5 shot\end{tabular}
& \begin{tabular}{c}0.714 \\0.846\end{tabular}\\
\noalign{\smallskip}\hline\noalign{\smallskip}
Split 3 &
\begin{tabular}{c}5 way 1 shot \\5 way 5 shot\end{tabular}
& \begin{tabular}{c}0.767 \\0.933\end{tabular}\\
\noalign{\smallskip}\hline
\end{tabular}
\label{tab:piwres}
\end{table}
\vspace{-1em}
\section{Conclusion}
\vspace{-5pt}
In this paper, we address the problem of few-shot learning in agriculture. Most of the current works in the field use instance embeddings that are not tailored to the task at hand. Using a Transformer based architecture \cite{vaswani2017attention, ye2020few}, we enrich the embeddings of the images by considering all the images in the support set of the given episode. We further incorporate a class-covariance-based deterministic metric - Mahalanobis distance \cite{galeano2015mahalanobis, bateni2020improved} to calculate the similarity of the query vector with the candidates of the support set. We show that our architecture significantly improves the performances on two different datasets (alongside one cross-dataset experiment) under multiple settings. 

We also provide a new dataset (\textit{Plants in Wild}) mimicking the real-world scenarios. Our newly compiled dataset consists of plants that are both healthy and diseased. Using our model which has been shown to outperform several existing FSL architectures on agriculture datasets, we provide strong baselines on our new dataset. We hope that our new dataset along with our new models opens up new research directions in the field of agriculture.

{\small
\bibliographystyle{ieee_fullname}
\bibliography{egbib}
}
\end{document}